\documentclass{article}

\PassOptionsToPackage{numbers, compress}{natbib}

\usepackage[final]{neurips_2023}

\usepackage[utf8]{inputenc} 
\usepackage[T1]{fontenc}    
\usepackage{hyperref}       
\usepackage{url}            
\usepackage{booktabs}       
\usepackage{amsfonts}       
\usepackage{nicefrac}       
\usepackage{microtype}      
\usepackage{graphicx}       
\usepackage{multirow}
\usepackage[table]{xcolor}  

\usepackage{enumitem}

\title{XGen-7B Technical Report}

\author{%
Erik Nijkamp\thanks{indicates lead authors, $^\dagger$indicates lead project coordinators and corresponding authors.},~ Tian Xie$^\ast$, Hiroaki Hayashi$^\ast$, Bo Pang$^\ast$, Congying Xia$^\ast$, Chen Xing \And  Jesse Vig,  Semih Yavuz, Philippe Laban, Ben Krause, Senthil Purushwalkam, Tong Niu \And  Wojciech Kry{\'s}ci{\'n}ski, Lidiya Murakhovs'ka, Prafulla Kumar Choubey, Alex Fabbri \And Ye Liu, Rui Meng,  Lifu Tu, Meghana Bhat,  Chien-Sheng Wu, Silvio Savarese \And Yingbo Zhou$^\dagger$, Shafiq Joty$^\dagger$, Caiming Xiong$^\dagger$ \\ \\
Salesforce Research
}

\usepackage{amsmath,amsfonts,bm}
\usepackage[nameinlink]{cleveref}

\crefformat{section}{\S#2#1#3} 
\crefname{algorithm}{Alg.}{Algs.}
\crefformat{subsection}{\S#2#1#3}
\Crefname{equation}{Eq.}{Eqs.}
\Crefname{figure}{Fig.}{Figs.}

\newcommand{\xgen}{{XGen-7B}}
\newcommand{\llamatwo}{{LLaMA-2-7B}}
\newcommand{\llama}{{LLaMA-7B}}
\newcommand{\openllama}{{OpenLLaMA-7B}}
\newcommand{\falcon}{{Falcon-7B}}
\newcommand{\mpt}{{MPT-7B}}
\newcommand{\opt}{{OPT-13B}}
\newcommand{\gptj}{{GPT-J-6B}}
\newcommand{\cerebras}{{Cerebras-GPT-13B}}
\newcommand{\redpajama}{{Redpajama-7B}}
\newcommand{\dolly}{{Dolly-v2-12B}}
\newcommand{\gptneo}{{GPT-neox-20B}}
\newcommand{\stablelm}{{StableLM-alpha-7B}}

\newcommand{\gptf}{{GPT-4}}
\newcommand{\claude}{{Claude}}
\newcommand{\chatgpt}{{ChatGPT}}
\newcommand{\wizardlm}{{WizardLM}}
\newcommand{\vicuna}{{Vicuna}}
\newcommand{\davincithr}{{text-davinci-003}}
\newcommand{\davincione}{{text-davinci-001}}
\newcommand{\alpaca}{{Alpaca}}
\newcommand{\falconinst}{{Falcon-40B-instruct}}
\newcommand{\guanaco}{{Guanaco-65B}}
\newcommand{\oasst}{{OAsst-RLHF-LLaMA-33B}}

\newcommand{\xgenwizard}{{XGen-7B-Inst}$_{\text{wizardLM}}$}
\newcommand{\xgenold}{{XGen-7B-Inst}$_{\text{general}}$}

\makeatletter   
\newcommand{\sveryshortarrow}[1][3pt]{\mathrel{%
    \vcenter{\hbox{\rule[-.5\fontdimen8\scriptfont3]
               {\scriptratio\dimexpr#1\relax}{\fontdimen8\scriptfont3}}}%
   \mkern-4mu\hbox{\let\f@size\sf@size\usefont{U}{lasy}{m}{n}\symbol{41}}}}
\makeatother

\def\eqref#1{equation~\ref{#1}}

\def\1{\bm{1}}

\def\m1{{\bm{1}}}

\DeclareMathAlphabet{\mathsfit}{\encodingdefault}{\sfdefault}{m}{sl}
\SetMathAlphabet{\mathsfit}{bold}{\encodingdefault}{\sfdefault}{bx}{n}

\begin{document}

\maketitle

\begin{abstract}
Large Language Models (LLMs) have become ubiquitous across various domains, transforming the way we interact with information and conduct research. However, most high-performing LLMs remain confined behind proprietary walls, hindering scientific progress. Most open-source LLMs, on the other hand, are limited in their ability to support longer sequence lengths, which is a key requirement for many tasks that require inference over an input context. To address this, we have trained \xgen, a series of 7B parameter models on up to 8K sequence length for up to 1.5T tokens. We have also finetuned the \xgen\ models on public-domain instructional data, creating their instruction-tuned counterparts (\xgen-Inst). We open-source our models for both research advancements and commercial applications. Our evaluation on standard benchmarks shows that \xgen\ models achieve comparable or better results when compared with state-of-the-art open-source LLMs. Our targeted evaluation on long sequence modeling tasks shows the benefits of our 8K-sequence models over 2K-sequence open-source LLMs.
\end{abstract}

\section{Introduction}

Large language models (LLMs) have shown impressive capabilities to generate text, translate languages, write code, answer questions, solve math problems, predict actions, and many more. Interestingly, they can perform these tasks from textual instructions and/or observing a few demonstrations \cite{brown2020language}. Crucial to their success are two main ingredients: (a) \emph{model scale} which defines the model’s capacity; and (b) \emph{instruction tuning}, which aims to align the model to follow user instructions \cite{instructgpt_ouyang2022training}. 

While the proliferation of LLMs has enhanced numerous applications, a significant number of high-performing models remain proprietary, impeding the progress of scientific exploration. Recent work \cite{hoffmann2022an} on model scaling has shown that for a given compute budget, the best performances are not necessarily achieved by the largest models, but by smaller models trained on more data (measured by the number of tokens). A smaller model is also generally preferred for inference efficiency during serving including on mobile devices.   

As LLMs become ubiquitous, their applications to long sequences have been a key focus \cite{tay2021long,shaham-etal-2022-scrolls}, especially for applications like writing code, summarizing text (potentially interleaved with other data sources like tables and images), and predicting protein sequences, which require the model to effectively consider long distance structural dependencies. A large context allows a pre-trained LLM to look at customer data (e.g., documents the LLM did not use in training) and responds to useful information seeking queries. Yet, most open-source LLMs (e.g., LLaMA \cite{touvron2023llama}, MPT\footnote{\url{https://www.mosaicml.com/blog/mpt-7b}}, Falcon\footnote{\url{https://falconllm.tii.ae/}}) have been trained with a maximum of 2K token sequence length, which is a key limitation in modeling long sequences. Inference time solutions such as ALiBi \cite{press2022train} have yet to be tested properly for larger models (e.g. MPT-7B-StoryWriter-65k+).

To address the above limitation, in light of the scaling properties and serving efficiency, we train a series of 7B LLMs named \xgen\ with standard dense attention on up to 8K sequence length for up to 1.5T tokens. We also finetune the \xgen\ models on public-domain instructional data, creating their instruction-tuned counterparts (\xgen-Inst). We open-source our models for both research advancements and commercial applications. Table \ref{tab:main} summarizes our released models.\footnote{\url{https://github.com/salesforce/XGen}}.

\begin{table}[tb]
\small 
  \center
  \caption{High-level summary of the \xgen\ models.}
  \begin{tabular} {ll}
  \toprule
  \bf Model & \bf Description \\  
  \midrule
\multirow{ 2}{*}{\xgen-4K} &  Pre-train for 800B tokens with a sequence length of 2K tokens first, \\ & then for another 400B tokens (total 1.2T tokens) with 4K tokens. \\
\multirow{ 2}{*}{\xgen-8K} & Initialize with \xgen-4K-base and further train for 300B more \\ & tokens (total 1.5T tokens) with 8K sequence length. \\
\midrule
\multirow{ 2}{*}{\xgenwizard} & Supervised fine-tuning of \xgen-8K on the recently released \\ & WizardLM-196K  \cite{xu2023wizardlm} instruction data.\\ 
\multirow{ 2}{*}{\xgenold} & Supervised fine-tuning of \xgen-8K on general public domain  instruction \\ & data including  OAsst\footnote{\url{https://huggingface.co/datasets/OpenAssistant/oasst1}}, Baize \cite{xu2023baize}, Dolly2 \cite{DatabricksBlog2023DollyV2}, ShareGPT and SCROLLS \cite{shaham-etal-2022-scrolls}.\\ 
  \bottomrule
  \end{tabular}
\label{tab:main}
\end{table}

Our evaluation of \xgen-8K on standard benchmarks for evaluating base pre-trained models shows that it achieves comparable or better results when compared with state-of-the-art open-source LLMs. It also achieves good results on Python code generation tasks. Our instruction-tuned models also show impressive results on the recently proposed AlpacaEval \cite{alpaca_eval} and MTBench \cite{zheng2023judging} benchmarks, often outperforming models of similar sizes (e.g., \wizardlm-7B, \mpt) and even larger ones (e.g., \falconinst, \alpaca-13B). Furthermore, our targeted evaluation on long sequence modeling tasks show benefits of our 8K-sequence models over 2K-sequence open-source LLMs.

\section{Pre-training Data}

Our pre-training dataset is a mixture of data from several public sources, reported in Table \ref{tab:dataset_1}. We employ a two-stage training strategy, where each stage uses a different data mixture, as shown in Table \ref{tab:dataset_2}. 

\textbf{Natural language data for stage 1.} Natural language data is a mixture of publicly available data. We made an effort to improve safety and diversity of the data.

\textbf{Code data for stage 1.} We use the GitHub subset from the recently released RedPajama dataset~\cite{together2023redpajama}. We also added Apex code data to enhance our model's proficiency in Apex code generation. Apex is a widely used object-oriented programming language in Salesforce products.

\textbf{BigCode Starcoder data for stage 2.} We use all the 86 programming languages from the Starcoder~\cite{li2023starcoder} data, preserving the original weight of each. Subsequently, we further filter the data according to a stronger permissive license guideline.

\begin{table}[ht]
  \center
  \caption{Data mixtures used for pre-training stage 1. For each subset of the data, we report the effective number of tokens, and its sampling proportion.
  \label{tab:dataset_1}
  }
  \setlength{\tabcolsep}{3pt}
  \begin{tabular} {lrrr}
  \toprule
  \bf Dataset &  \bf Tokens (B) & \bf Sampling prop. (\%) \\  %
  \midrule
  Natural language data & \phantom{0}1309.99 & 95.31 \\
  Code data & \phantom{00}64.53  & \phantom{0}4.69 \\
  \midrule
  Total &  1374.52 & 100 \\
  \bottomrule
  \end{tabular}
\end{table}

\begin{table}[t]
  \center
  \caption{Data mixtures used for pre-training stage 2. 
  \label{tab:dataset_2}
  }
  \setlength{\tabcolsep}{3pt}
  \begin{tabular} {lrr}
  \toprule
  \bf Dataset &  \bf Tokens (B) &  \bf Sampling prop. (\%) \\  %
  \midrule
  Data from stage 1    & \phantom{0}55 & \phantom{0}50 \\
  BigCode Starcoder data     & \phantom{0}55 & \phantom{0}50 \\
  \midrule
  Total &  110  & 100 \\
  \bottomrule
  \end{tabular}
\end{table}

\textbf{Tokenizer.} We tokenize the data with the byte pair encoding (BPE) algorithm~\cite{sennrich2015neural}, utilizing OpenAI's tiktoken tool, with GPT-2 serving as the base tokenizer. Additionally, we incorporate supplementary special tokens as outlined in the Starcoder paper \cite{li2023starcoder}, along with consecutive white-spaces and tabs with the goal of aiding code generation.

\textbf{Constructing sequences of different lengths.} During the pre-training stage 1, there are 3 substages, each with varying sequence lengths: 2K, 4K, and 8K tokens. To ensure data integrity and prevent potential distributional shifts, we shuffle the data uniformly, and split the shuffled data into 3 big chunks for the 3 substages. We construct the training sequences by concatenating or splitting the original text documents into the target sequence lengths. When two different documents are concatenated, an \texttt{<|endoftext|>} token is added between them. We exclude short documents that contain less than 100 tokens after tokenization. We then shuffle the constructed training sequences uniformly in each big chunk.
The data for pre-training stage 2 (50\% stage 1 data and 50\% Starcoder data) only has training sequences with a length of 8k tokens.

\section{Training Details}

\begin{figure}[tb]
  \centering
  \includegraphics[width=0.6\textwidth]{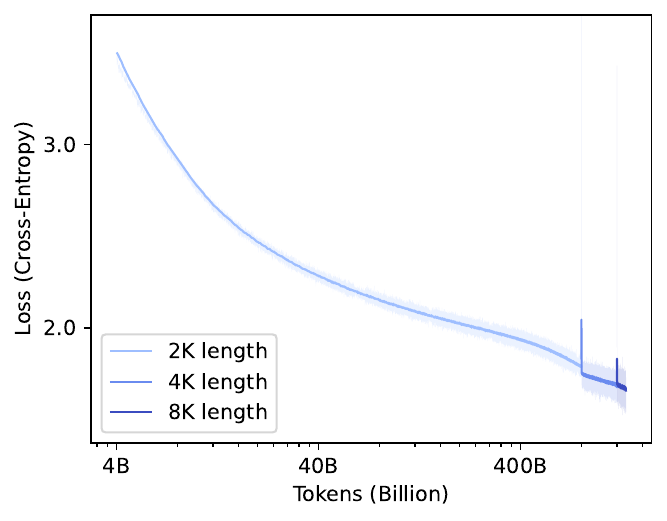}
  \caption{Cross-entropy over training time. The model is pre-trained in three stages with a step-wise increase of the sequence length from 2K to 4K to 8K tokens. Note, the pre-training does not suffer from any loss spikes. The spikes at the transitions in sequence lengths are expected as the model adjusts to positional encodings of increased length. The drop in perplexity from 2K to 4K is expected as uncertainty decreases over sequence length for long sequences.}
  \label{fig:train_loss}
\end{figure}

The \xgen\ models are trained with our library JaxFormer~\cite{Jaxformer}, which facilitates efficient training of LLMs under both data and model parallelism optimized for TPU-v4 hardware.
The training recipe and model architecture follow LLaMA~\cite{touvron2023llama}, while we conduct two additional explorations. First, we investigate the occurrence of so-called ``loss spikes''~\cite{chowdhery2022palm,gpt-j,molybog2023theory} during training, that is, the loss suddenly explodes temporarily while the root cause for these spikes is unknown. Second, the \xgen\ models support sequence lengths of up to 8,192 tokens (rather than the common 2,048) for which we introduce stage-wise training.

\begin{figure}[bt]
  \centering
  \includegraphics[width=0.6\textwidth]{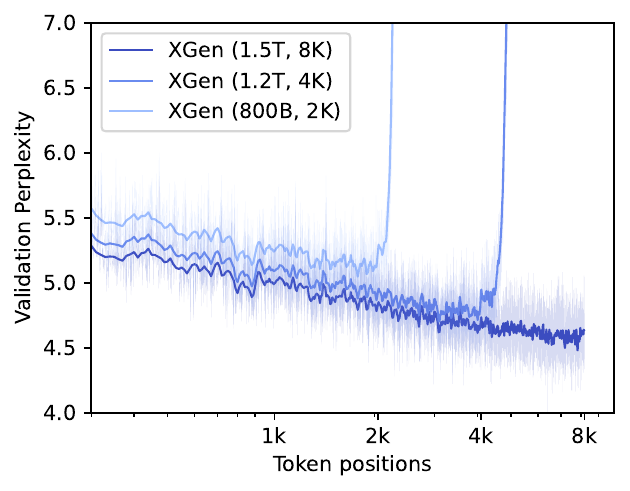}
  \caption{Perplexity over sequence length. If a model can utilize the information encoded in long sequences, then, in expectation, the perplexity should decrease over the length of such sequences. That is, the information contained in previous tokens increases the certainty of the next token prediction, which can be observed in the figure. Note, the perplexity of the 8K over the 2K model is generally lower as the model has been trained for an additional 700B tokens.}
  \label{fig:train_ppl}
\end{figure}

\paragraph{Recipe.} The model architecture follows LLaMA with exact numerical compatibility to ease adoption in third-party frameworks. The hyperparameters closely follow LLaMA-7B~\cite{touvron2023llama} with the following alterations: (1) The token budget has been increased from 1.0T to 1.5T tokens, (2) the training is performed stage-wise to increase the sequence length from 2K to 4K to 8K, (3) the vocabulary size has been increased from 32,000 to 51,200 tokens. The training loop was implemented in JAX with Haiku for which the entire computation is under FP32 numerical precision, except for matmul in BF16.

\paragraph{Loss spikes.} As models are scaled to larger sizes, the training itself is increasingly sensitive to instabilities, which cause poor model performance, if not addressed carefully~\cite{chowdhery2022palm,gpt-j,molybog2023theory}. In our exploration, we have gathered evidence for several factors, which individually contribute to unstable training. These preliminary findings include ``sequential over parallel self-attention circuits'', ``swish-GLU over GeLU'', ``RMS-Norm over Layer-norm''. Specifically, widely used parallel circuits~\cite{chowdhery2022palm,gpt-j,nijkamp2022codegen}, which parallelize the computation of self-attention and feed-forward may affect the stability of training, at least in our specific setting. As adopted in~\cite{touvron2023llama}, the combination of activation normalization in the form of RMS-Norm~\cite{zhang2019root}, sequential self-attention and swish-GLU~\cite{shazeer2020glu} appears to be numerically highly robust, while not optimal in terms of computational efficiency.

\paragraph{Sequence length.} Training with longer sequences is computationally unproportionally costly as the complexity of self-attention is quadratic, that is, the training process is slow. To mitigate slow training, we introduce training in stages with increasing sequence length. First, 800B tokens with sequence length of 2K tokens are observed, then 400B tokens with 4K, finally, 300B tokens with 8K length. Figure~\ref{fig:train_loss} shows the cross-entropy over training steps for this stage-wise training. We verify the adaptation to longer sequences by computing the average perplexity at each token position on a held-out validation set containing documents of 8K sequence length or above. If the model successfully learns to utilize the full sequence, we would expect the perplexity to decrease over sequence length, as previous tokens carry information for the next to-be-predicted token. That is, for a long sentence, the more context in the form of previous words is provided, the easier it becomes to guess the next word. Figure~\ref{fig:train_ppl} indeed demonstrates that \xgen\ at each stage successfully learns to utilize longer contexts, up to 8K sequence length.

\section{Instructional Tuning}

To demonstrate the language understanding and generation capability of \xgen, we perform instruction-tuning of the base LLM and evaluate the instruct-tuned models.

\textbf{Instruction Data.}
The key to instruction tuning is the \emph{instructional data} that is used to align the model to follow user instructions, while being harmless. Developing proprietary models like GPT-4~\cite{openai2023gpt4} and Bard~\cite{nicholson2020bard} involves significant annotation efforts for collecting such data. Early open-source instruction tuned models~\cite{longpre2023flan} leverage academic datasets by transforming them into instructional formats with human-written prompt templates~\cite{wei2021finetuned, sanh2021multitask, supernaturalinstructions}. Despite the large amount of data curated in this way, instruction following capacity of these models falls behind proprietary models as the task distributions covered by these academic benchmarks do not match the real use cases of LLMs \cite{instructgpt_ouyang2022training}.

Therefore, more recent open-source  models~\cite{zheng2023judging, xu2023wizardlm} utilize ChatGPT- or GPT4-synthesized data, e.g., human-written prompts with GPT-generated responses or GPT-generated prompts and responses. This distillation process helps to close the gap to the proprietary models. Some examples of these datasets are Alpaca \cite{alpaca}, ShareGPT~\footnote{\url{https://huggingface.co/datasets/anon8231489123/ShareGPT_Vicuna_unfiltered}}, Baize \cite{xu2023baize}, GPTeacher~\footnote{\url{https://github.com/teknium1/GPTeacher}}, and WizardLM \cite{xu2023wizardlm}.

For our experiments, we finetune \xgen\ in two data settings:

\begin{itemize}[leftmargin=*]
    \item \xgenwizard: For this setting, we use WizardLM \cite{xu2023wizardlm}, which is one of the most recent instruction datasets. It is created by prompting GPT-4 to rewrite existing instructions from Alpaca \cite{alpaca} to make them more complex. Finetuning LLaMA models on this dataset has demonstrated high performance in several benchmarks especially for complex instructions. We use the WizardLM-196K collection for finetuning \xgen. This setting allows us to compare with the WizardLM-7b model \cite{xu2023wizardlm}, which is based on LLaMA-7b and uses the same source of instructions. 

    \item \xgenold: In another setting, we use general public domain instruction data that includes OAsst\footnote{\url{https://huggingface.co/datasets/OpenAssistant/oasst1}}, Baize \cite{xu2023baize}, Dolly2 \cite{DatabricksBlog2023DollyV2} and ShareGPT. To measure the impact of long contexts, we also include examples from the long sequence NLP benchmark, SCROLLS \cite{shaham-etal-2022-scrolls}. We sample about 1,500 examples from each of the following datasets from SCROLLS: GovReport, SummScreenFD, QMSum, NarrativeQA, Qasper and QuALITY. We sample these examples such that each contains at least 4000 tokens. This setting is intended to give us a sense on the model's general capability in following instructions for long-sequence tasks. 
    
\end{itemize}

 \textbf{Finetuning Details.} We use Adam with $\beta_1=0.9$ and $\beta_2=0.99$, cosine decay for learning rate down to 10\% of an initial value $2 \times 10^{-5}$, a batch size of 128, and a sequence length of 8,192 tokens. Each data instance is formatted as a single-turn or multi-turn conversation between Human and Assistant. In particular, it follows the format:
 
 \texttt{\#\#\# Human: \{prompt\} \#\#\# Assistant: \{response\}}

Our training objective is causal language modeling and the loss for \texttt{prompt} is masked out, thus only the gradients for \texttt{response} tokens are backpropagated. We train our models for 3 epochs.  

\section{Evaluation}

\subsection{Base Model Evaluation}

\subsubsection{Standard NLP Benchmarks}

We first consider Massive Multitask Language Understanding benchmark \cite{hendrycks2020measuring}, which is more recent and less susceptible to data contamination as reported in recent studies (see page 32 of GPT-4 technical report \cite{openai2023gpt4} and a related discussion \footnote{https://hitz-zentroa.github.io/lm-contamination/blog/?ref=blog.salesforceairesearch.com}). The benchmark has been widely adopted for held-out evaluation. Recently, however, inconsistencies in reporting MMLU scores have been reported, which resulted in wrong rankings in Hugging Face’s Open LLM leaderboard. In our work, we follow the original MMLU standard, which is consistent with the published results (i.e., in LLaMA).

MMLU consists of multiple choice questions covering various domains of knowledge, including humanities, STEM and social sciences. To assess our models' performance, we conduct evaluations in both the five- and zero-shot settings, utilizing the sample questions from the benchmark. The results for five-shot MMLU are reported in Table \ref{mmlu_5}, and the results for zero-shot MMLU are reported in Table \ref{mmlu_0}. For both settings, {\xgen} achieves the best results among the baselines in most categories, as well as in weighted average.

\begin{table}[tb]
\centering
\caption{Five-shot results (accuracy) on Massive Multitask Language Understanding (MMLU).}
\resizebox{0.9\linewidth}{!}{
\begin{tabular}{lccccc}

\toprule
{ \textbf{Models}}           & { \textbf{Humanities}} & { \textbf{STEM}} & { \textbf{Social Sciences}} & { \textbf{Other}} & { \textbf{Weighted average}} \\ \midrule
{ \xgen}          & { 33.8}                & { 30.7}          & { 40.0}                     & { 41.5}           & { 36.3}                      \\
 
{ \llama}         & { 33.9}                & { 30.6}          & { 38.2}                     & { 38.2}           & { 35.1}                      \\
 
{ \openllama}     & { 28.1}                & { 28.5}          & { 31.2}                     & { 32.8}           & { 29.9}                      \\
 { \falcon }        & { 26.5}                & { 25.4}          & { 29.2}                     & { 26.8}           & { 26.9}                      \\
 { \mpt }           & { 25.9}                & { 26.2}          & { 26.9}                     & { 28.1}           & { 26.7}                      \\
 
{ \redpajama}     & { 26.1}                & { 25.2}          & { 27.4}                     & { 26.7}           & { 26.3}                      \\
 
{ \cerebras} & { 26.1}                & { 26.5}          & { 25.8}                     & { 26.6}           & { 26.2}                      \\
 
{ \dolly}     & { 26.9}                & { 25.7}          & { 25.3}                     & { 26.5}           & { 26.2}                      \\
 
{ \opt}          & { 26.2}                & { 24.3}          & { 23.4}                     & { 26.0}             & { 25.1}                      \\
 
{ \gptj}         & { 25.9}                & { 24.0}          & { 24.0}                     & { 25.8}           & { 25.1} \\ \bottomrule
\end{tabular}
}
\label{mmlu_5}
\end{table}

\begin{table}[tb]
\centering
\caption{Zero-shot accuracy on Massive Multitask Language Understanding (MMLU).}
\resizebox{0.9\linewidth}{!}{
\begin{tabular}{lccccc}
 \toprule
{ \textbf{Models}}           & { \textbf{Humanities}} & { \textbf{STEM}} & { \textbf{Social Sciences}} & { \textbf{Other}} & { \textbf{Weighted average}} \\ \midrule
{ \xgen}          & { 31.4}                & { 27.8}          & { 32.1}                     & { 37.2}           & { 32.1}                      \\
 
{ \llama}         & { 32.3}                & { 27.1}          & { 31.3}                     & { 36.8}           & { 32.0}                      \\
 
{ \openllama}     & { 28.0}                & { 27.6}          & { 28.9}                     & { 30.1}           & { 28.6}                      \\
 
{ \mpt}           & { 27.4}                & { 25.2}          & { 26.0}                     & { 30.7}           & { 27.4}                      \\
 
{ \redpajama}     & { 27.5}                & { 25.5}          & { 24.2}                     & { 25.0}           & { 25.8}                      \\
 
{ \gptj}         & { 25.3}                & { 24.5}          & { 25.5}                     & { 27.6}           & { 25.7}                      \\
 
{ \dolly}     & { 26.2}                & { 26.0}          & { 24.0}                     & { 24.9}           & { 25.4}                      \\
 
{ \cerebras} & { 24.3}                & { 25.0}          & { 23.0}                     & { 26.0}           & { 24.6}                      \\
 
{ \opt}          & { 26.3}                & { 23.3}          & { 23.6}                     & { 23.6}           & { 24.4}                      \\
 
{ \falcon }        & { 24.8}                & { 21.7}          & { 24.0}                     & { 24.4}           & { 23.9}        \\ \bottomrule          
\end{tabular}
}
\label{mmlu_0}
\end{table}

\begin{table}[tb]
\caption{Zero-shot performance on Common Sense Reasoning and Question Answering tasks.}
\resizebox{\linewidth}{!}{
\begin{tabular}{lcccccccc}
 \toprule
{ \textbf{Models}}            & { \textbf{\begin{tabular}[c]{@{}c@{}}MMLU\\ -wavg\end{tabular}}} & { \textbf{ARC\_ch}} & { \textbf{HellaSwag}} & { \textbf{Winogrande}} & { \textbf{TruthfulQA}} & \multicolumn{1}{l}{\cellcolor[HTML]{FFFFFF}{ \textbf{BoolQ}}} & { \textbf{PiQA}} & { \textbf{OpenBookQA}} \\ \midrule
{ \xgen}           & { 32.1}                                                          & { 41.2}             & { 74.2}                & { 64.9}                & { 39.1}                & { 74.3}                                                       & { 75.5}          & { 40.2}                \\
 
{ \llama}          & { 32.0}                                                          & { 44.8}             & { 76.2}                & { 69.6}                & { 34.0}                  & { 74.9}                                                       & { 78.7}          & { 44.2}                \\
 
{ \falcon}         & { 23.9}                                                          & { 43.4}             & { 76.4}                & { 67.2}                & { 34.3}                & { 73.8}                                                       & { 79.4}          & { 44.0}                \\
 
{ \mpt}            & { 27.4}                                                          & { 41.7}             & { 76.1}                & { 68.6}                & { 33.4}                & { 74.1}                                                       & { 79.1}          & { 41.8}                \\
 
{ \openllama}      & { 28.6}                                                          & { 38.7}             & { 71.8}                & { 67.0}                & { 35.2}                & { 70.6}                                                       & { 76.0}          & { 39.0}                \\
 
{ \redpajama}      & { 25.8}                                                          & { 39.1}             & { 70.3}                & { 63.8}                & { 33.3}                & { 69.3}                                                       & { 76.9}          & { 40.0}                \\
 
{ \gptneo}      & { 24.5}                                                          & { 41.1}             & { 70.5}                & { 66.1}                & { 31.4}                & { 64.9}                                                       & { 76.7}          & { 38.8}                \\
 
{ \opt}           & { 24.4}                                                          & { 35.8}             & { 69.9}                & { 64.7}                & { 33.9}                & { 65.0}                                                       & { 75.7}          & { 39.8}                \\
 
{ \gptj}          & { 25.7}                                                          & { 36.3}             & { 66.2}                & { 64.5}                & { 36.0}                & { 65.4}                                                       & { 75.4}          & { 38.2}                \\
 
{ \dolly}      & { 25.4}                                                          & { 39.6}             & { 70.8}                & { 61.8}                & { 34.4}                & { 56.3}                                                       & { 75.4}          & { 39.2}                \\
 
{ \cerebras}  & { 24.6}                                                          & { 32.4}             & { 59.4}                & { 60.8}                & { 39.2}                & { 61.1}                                                       & { 73.5}          & { 35.8}                \\
 
{ \stablelm} & { 24.4}                                                          & { 27.0}             & { 40.7}                & { 51.5}                & { 41.7}                & { 59.0}                                                       & { 65.8}          & { 32.4}               \\ \bottomrule
\end{tabular}
}
\label{nlp_zeroshot}
\end{table}

We also report general zero-shot results on other standard NLP benchmarks that involve common sense reasoning and QA: ARC challenge \cite{clark2018think}, HellaSwag \cite{zellers2019hellaswag}, Winogrande \cite{sakaguchi2021winogrande}, TruthfulQA \cite{lin2021truthfulqa}, BoolQ \cite{clark2019boolq}, PiQA \cite{bisk2020piqa}, and OpenBookQA \cite{mihaylov2018can}. The datasets comprise Cloze and Winograd style tasks, alongside multiple-choice question answering. Our evaluation follows the zero-shot approach commonly employed in the language modeling community \cite{eval-harness}. As shown in Table \ref{nlp_zeroshot}, {\xgen} achieves comparable performance to the state-of-the-art LLMs of similar sizes.

\subsubsection{Code Generation}

To evaluate \xgen’s code generation capability from natural language instructions (i.e., docstrings), we evaluate the model on HumanEval benchmark~\cite{chen2021codex}. HumanEval evaluates LLMs' Python code-writing capabilities at the function level by assessing functional correctness. We report performance using the pass@1 metric~\cite{chen2021codex}. A generated code is considered correct if it passes all the unit tests. Following~\cite{chen2021codex}, we set the sampling temperature to $0.2$, $p = 0.95$ for top-$p$ sampling, and generate $n = 200$ samples for each problem in the benchmark to report an unbiased pass@1 score. As we can notice in Table \ref{tab:HumanEval}, our \xgen\ achieves comparable results to state-of-the-art 7B LLMs. 

\begin{table}[tb]
  \center
  \caption{Natural language to code generation results in pass@1 on the HumanEval benchmark. For \openllama-v2, we note that Starcoder data occupies 30\% of their pre-training data. $^\ast$Consecutive whitespaces are treated as one, breaking Python syntax. $^{\ast\ast}$Model could not generate meaningful code.}
  \setlength{\tabcolsep}{3pt}
  \begin{tabular} {lc}
  \toprule
  \bf Models &  \bf pass@1 \\  %
  \midrule
  \xgen & 14.20 \\
  \mpt  & 15.90 \\
  \openllama-v2 & 14.83 \\
  \llamatwo & 13.55 \\
  \llama    & 10.38 \\
  \redpajama & \phantom{0}5.24 \\
  \openllama & 0$^\ast$\phantom{$^\ast$} \\
  \falcon & 0$^{\ast\ast}$ \\
  \bottomrule
  \end{tabular}
  \label{tab:HumanEval}
\end{table}

Considering the size and results in both text and code tasks, \xgen\ can be a good general-purpose model that can be served both on standard-sized GPUs (e.g., 16 GB memory) and mobile devices.

\subsection{Instruction Model Evaluation}

\subsubsection{AlpacaEval}

AlpacaEval\footnote{\url{https://tatsu-lab.github.io/alpaca_eval/}} \cite{alpaca_eval} is a newly proposed automated evaluation platform that employs an LLM as an evaluator. It utilizes the AlpacaFarm \cite{dubois2023alpacafarm} evaluation dataset, which has been crafted to evaluate a model's ability to understand and follow a wide range of user instructions. The responses generated by the models under evaluation are then contrasted with the reference responses from \davincithr\ \citep{instructgpt_ouyang2022training}, with GPT-4 \cite{openai2023gpt4} serving as the evaluator. The win rate against \davincithr is employed as the performance metric. 

\begin{table}[tb]
\centering
\caption{Results on the AlpacaEval leaderboard \cite{alpaca_eval} with GPT-4 as an evaluator.}
\begin{tabular}{lc} \toprule
\textbf{Model}                                                                    & \multicolumn{1}{l}{\textbf{Win Rate vs. text-davinci-003}} \\ \midrule
\gptf                                                                             & 95.3                                                \\
\claude                                                                            & 88.4                                                \\
\chatgpt                                                                           & 86.1                                                \\
\vicuna-7B-V1.3                                                                   & 76.8                                                \\
\wizardlm-13B                                                                      & 75.3                                                \\
\guanaco                                                                       & 71.8                                                \\
\vicuna-13B                                                                        & 70.4                                                \\
\rowcolor[gray]{0.85}
\xgenwizard                                                            & 68.8                                                \\
\wizardlm-7B                                                                      & 65.2                                                \\
\oasst                                                              & 66.5                                                \\
\vicuna-7B                                                                        & 64.4                                                \\
\rowcolor[gray]{0.85}
\xgenold                                                    & 57.3                                                \\
\davincithr                                                                & 50.0                                                  \\
\falconinst                                                              & 45.7                                                \\
\mpt-chat                                                                       & 45.0                                                  \\
Alpaca-farm-PPO-human                                                             & 41.2                                                \\
\alpaca-7B                                                                         & 26.5                                                \\
\davincione                                                               & 15.2                                       \\ \bottomrule        
\end{tabular}
\label{alpaca_eval}
\end{table}

As shown in Table \ref{alpaca_eval}, our instruction-tuned model, \xgenwizard\ (fine-tuned on WizardLM \cite{xu2023wizardlm}), generally achieves better performance than other models of similar sizes, notably the \wizardlm-7B, which uses the same repository of distilled instructions. Our model performs slightly worse than Vicuna-7B-v1.3, which utilizes more ShareGPT data comprising human-authored prompts. The \xgenold\ model performs worse than \xgenwizard\ but still significantly better than \davincithr\ and other open-source alternatives like \falconinst\ and MPT-7B-chat. 

\subsubsection{MT-Bench}

Similar to AlpacaEval, MT-Bench\footnote{\url{https://huggingface.co/spaces/lmsys/mt-bench}} \cite{zheng2023judging} is a new benchmark for evaluating LLM-based chat assistants. It also uses an LLM as a judge (e.g., GPT-4) to assess the models on open-ended questions. The model evaluation is performed in two ways:

\textbf{Single answer grading.} In this evaluation 
 setting, the judge LLM assigns a score directly to each of the model-generated responses. As shown in Table \ref{tab:mt_single}, \xgenwizard\ outperforms other models of similar sizes (except Vicuna-7B-v1.3), especially the \wizardlm-7B-Inst model which uses a similar instruction set. It even surpasses larger models, such as the Falcon-40B-instruct or MPT-30B-instruct.

\begin{table}[ht]
\small
\centering
\caption{Evaluation on MT Bench \cite{zheng2023judging} -- Single answer grading by GPT-4.}
\begin{tabular}{lc}
\toprule
\textbf{Model} & \textbf{Score} \\ \midrule
 \gptf & 8.99 \\
\chatgpt (GPT-3.5-turbo) & 7.94 \\
\claude-v1 & 7.90 \\
\claude-instant-v1 & 7.85 \\ \midrule
\vicuna-33B-v1.3 & 7.12 \\
\wizardlm-30B & 7.01 \\
Guanaco-33B & 6.53 \\
Tulu-30B & 6.43 \\
Guanaco-65B & 6.41 \\
OAsst-SFT-7-LLaMA-30B & 6.41 \\
PaLM-2-chat-bison-001 & 6.40 \\
MPT-30B-chat & 6.39 \\
\vicuna-13B-v1.3 & 6.39 \\
\wizardlm-13B & 6.35 \\
\vicuna-7B-v1.3 & 6.00 \\
Baize-v2-13B & 5.75\\
\rowcolor[gray]{0.85} \xgenwizard & 5.69 \\
\rowcolor[gray]{0.85} \xgenold & 5.54 \\
Nous-Hermes-13B & 5.51 \\
MPT-7B-chat & 5.42 \\
GPT4All-13B-snoozy & 5.41 \\
Koala-13B & 5.35 \\
WizardLM-7B & 5.29 \\
MPT-30B-instruct & 5.22 \\
\falconinst & 5.17 \\
H2OGPT-OAsst-Open-LLaMA-13B & 4.63 \\
Alpaca-13B & 4.53 \\
ChatGLM-6B & 4.50 \\
OAsst-SFT-4-pythia-12B & 4.32 \\
RWKV-4-raven-14B & 3.98 \\
Dolly-v2-12B & 3.28 \\
Fastchat-T5-3B & 3.04 \\
StableLM-tuned-alpha-7B & 2.75 \\
LLaMA-13B & 2.61 \\
\bottomrule
\end{tabular}
\label{tab:mt_single}
\end{table}


\textbf{Pairwise comparison.} In this setting, the judge LLM is given a question along with two model responses from two competing models. The judge is tasked to determine which answer is superior, or to declare that both answers are equally good. From the results in Table \ref{tab:mt_pairwise}, we see that here also \xgen-Inst models outperform other models of similar sizes and they surpass some larger models.

\begin{table}
\centering
\caption{MT Bench Evaluation \cite{zheng2023judging} -- Pairwise Comparison by GPT-4.}
\scalebox{0.9}{
\begin{tabular}{lcccccc} 
\toprule
\textbf{Model} & \textbf{Win} & \textbf{Loss} & \textbf{Tie} & \textbf{Win Rate} & \textbf{Loss Rate} & \textbf{Win Rate Adjusted} \\ \midrule
\gptf & 111 & 7 & 42 & 69.4 & 43.8 & 82.5 \\
\claude-v1 & 75 & 27 & 58 & 46.9 & 16.9 & 65.0 \\
\vicuna-33B-v1.3 & 70 & 42 & 48 & 43.8 & 26.3 & 58.8 \\
\claude-instant-v1 & 64 & 40 & 56 & 40.0 & 25.0 & 57.5 \\ 
\wizardlm-30B & 37 & 63 & 60 & 23.1 & 39.4 & 41.9 \\
Guanaco-33B & 42 & 72 & 46 & 26.3 & 45.0 & 40.6 \\
Guanaco-65B & 38 & 68 & 54 & 23.8 & 42.5 & 40.6 \\
\vicuna-13B-v1.3 & 33 & 73 & 54 & 20.6 & 45.6 & 37.5 \\
MPT-30B-chat & 29 & 78 & 53 & 18.1 & 48.8 & 34.7 \\
\vicuna-7B-v1.3 & 60 & 165 & 95 & 18.8 & 51.6 & 33.6 \\
\wizardlm-13B & 27 & 81 & 52 & 16.9 & 50.6 & 33.1 \\
Tulu-30B & 29 & 92 & 39 & 18.1 & 57.5 & 30.3 \\
OAsst-SFT-7-LLaMA-30B & 23 & 88 & 49 & 14.4 & 55.0 & 29.7 \\
\rowcolor[gray]{0.85} \xgenwizard & 22 & 91 & 47 & 13.8 & 56.9 & 28.4 \\
Baize-v2-13B & 21 & 101 & 38 & 13.1 & 63.1 & 25.0 \\
PaLM-2-chat-bison-001 & 18 & 102 & 40 & 11.3 & 63.8 & 23.8 \\
\rowcolor[gray]{0.85} \xgenold & 17  & 108 & 35 & 10.6 & 67.5 & 21.6\\
Nous-Hermes-13B & 12 & 104 & 44 & \phantom{0}7.5 & 65.0 & 21.3 \\
GPT4All-13B-snoozy & 14 & 108 & 38 & \phantom{0}8.8 & 67.5 & 20.6 \\
MPT-7B-chat & 18 & 214 & 88 & \phantom{0}5.6 & 66.9 & 19.4 \\
H2OGPT-OAsst-Open-LLaMA-13B & 19 & 118 & 23 & 11.9 & 73.8 & 19.1 \\
Koala-13B & 10 & 110 & 40 & \phantom{0}6.3 & 68.8 & 18.8 \\
Falcon-40B-instruct & 10 & 116 & 34 & \phantom{0}6.3 & 72.5 & 16.9 \\
MPT-30B-instruct & 7 & 120 & 33 & \phantom{0}4.4 & 75.0 & 14.7 \\
ChatGLM-6B & 6 & 124 & 30 & \phantom{0}3.8 & 77.5 & 13.1 \\
OAsst-SFT-4-pythia-12B & 8 & 128 & 24 & \phantom{0}5.0 & 80.0 & 12.5 \\
RWKV-4-raven-14B & 6 & 128 & 26 & \phantom{0}3.8 & 80.0 & 11.9 \\
Alpaca-13B & 13 & 265 & 42 & \phantom{0}4.1 & 82.8 & 10.6 \\
Fastchat-T5-3B & 5 & 132 & 23 & \phantom{0}3.1 & 82.5 & 10.3 \\
Dolly-v2-12B & 5 & 138 & 17 & \phantom{0}3.1 & 86.3 & \phantom{0}8.4 \\
\bottomrule
\end{tabular}}
\label{tab:mt_pairwise}
\end{table}

\subsection{Long Sequence Tasks}

In addition to public benchmarks AlpacaEval and MT-Bench, we also evaluate \xgen\ and other competitive open source models on long sequence modeling tasks. 

\subsubsection{Long-form QA} \label{subsec:lqa}

In order to evaluate the reasoning capabilities of open source LLMs on long context, we design a long-form QA task in-house with two settings. Given a long input document, (1) we first prompt ChatGPT (GPT-3.5-turbo) to generate questions with explicit instructions such that answers are not directly retrievable from the context with few words. We call this setting \textbf{QG-passage}. (2) In order to generate more abstract questions that would require synthesizing different elements from different parts of the input document, we first summarize the document and then generate questions on the summary using ChatGPT similar to (1). We call this setting \textbf{QG-summary}. {We provide examples of the prompts in Appendix \ref{app:question_generation}}. 

Next, we prompt the models to answer the questions generated from ChatGPT on the two settings mentioned above. Note that we know the ground-truth answers in the two settings. We set a maximum of 512 tokens for generation. We use GPT-4 for evaluating the responses on the generated answers and rate them on a scale of 0-3 for the following dimensions: coherence, relevance, and accuracy. 

As shown in Table \ref{tab:long_form}, we find that \xgenold\  outperforms all the other models compared. Specifically, we find that the rates for \xgen-Inst models are higher for generated responses in terms of coherence and relevance. In general, we find that questions generated from summary are often more difficult to generate response which shows the difficulty of the overall setting (Table \ref{tab:cmp_longform}). These improvements can be partially attributed to XGen's long-sequence modeling capability. 

\begin{table}[tb]
\centering
\caption{{Overall performance} of different models based on GPT-4 evaluation on long-form QA. The table shows individual and average ratings across all metrics: coherence, relevance and accuracy.} 
\begin{tabular}{lccccc}
\toprule
\textbf{Model} & \textbf{Coherence} & \textbf{Relevance} & \textbf{Accuracy} & \textbf{Avg.} \\
\midrule
\xgenold & \textbf{2.81} & \textbf{2.72} & \textbf{2.70} & \textbf{2.74} \\
\vicuna-7B-v1.3 & 2.77 & 2.64 & 2.58 & 2.66 \\ 
\xgenwizard & 2.78 & 2.68 & 2.50 & 2.65 \\
\wizardlm-7B & 2.79 & 2.74 & 2.40 & 2.63 \\
\mpt-instruct & 2.55 & 2.48 & 2.30 & 2.43 \\
\falcon-instruct & 2.28 & 2.22 & 1.75 & 2.08 \\
Alpaca-7B & 1.65 & 1.91 & 1.58 & 1.71 \\
\bottomrule
\end{tabular}
\label{tab:long_form}
\end{table}

\begin{table}[tb]
\centering
\caption{Performance breakdown of different models in the two settings based on GPT-4 evaluation. The table shows average ratings across all metrics for questions generated from passage (QG-passage) and summary (QG-summary). 
}
\begin{tabular}{lccc}
\toprule
\textbf{Model} & \textbf{QG-passage} & \textbf{QG-summary} \\
\midrule
\xgenold & \textbf{2.79} & \textbf{2.68} \\ 
\vicuna-7B-v1.3 & 2.71 & 2.61 \\ 
\xgenwizard & 2.71 & 2.60 \\
\wizardlm-7B & 2.71 & 2.55 \\
\mpt-instruct & 2.50 & 2.35 \\
\falcon-instruct &  2.22 & 1.95 \\
Alpaca-7B &  2.04 & 1.64 \\
\bottomrule
\end{tabular}
\label{tab:cmp_longform}
\end{table}

\subsubsection{Dialogue Summarization}

In order to evaluate the long dialogue understanding and summarization capabilities, we perform experiments on three dialogue summarization tasks: AMI meeting summarization~\citep{AMI}, screenplay summarization from ForeverDreaming (FD) and TVMegaSite (TMS) datasets~\citep{chen-etal-2022-summscreen}. The average source lengths for these datasets are 5570, 6466, and 7653 tokens, respectively. 

For evaluation shown in Table~\ref{tab:long_dialogue}, we focus on samples with lengths less than 8K and consider the same instruction-tuned models as above. It is worth noting that both MPT-7B-inst and Alpaca-7B models performed poorly in this setting when input truncation was not applied. In contrast, our model (\xgen-Inst) achieved the highest ROUGE scores across all metrics.

\begin{table}[tb]
\centering
\caption{ROUGE scores of different models on long dialogue summarization task. 
\label{tab:long_dialogue}
}
\begin{tabular}{lccccccccc}
\toprule
\multirow{2}{5em}{\textbf{Model}} &
  \multicolumn{3}{c}{\textbf{AMI}} &
  \multicolumn{3}{c}{\textbf{FD}} &
  \multicolumn{3}{c}{\textbf{TMS}} \\ \cmidrule{2-4}\cmidrule{5-7}\cmidrule{8-10}
 &
  \multicolumn{1}{c}{R-1} &
  \multicolumn{1}{c}{R-2} &
  R-L &
  \multicolumn{1}{c}{R-1} &
  \multicolumn{1}{c}{R-2} &
  R-L &
  \multicolumn{1}{c}{R-1} &
  \multicolumn{1}{c}{R-2} &
  R-L \\ 
  \midrule
\xgenold &
  \multicolumn{1}{c}{\textbf{31.34}} &
  \multicolumn{1}{c}{\textbf{8.25}} &
  \textbf{17.00} &
  \multicolumn{1}{c}{\textbf{29.34}} &
  \multicolumn{1}{c}{\textbf{5.39}} &
  \textbf{16.43} &
  \multicolumn{1}{c}{\textbf{26.39}} &
  \multicolumn{1}{c}{\textbf{3.94}} &
  \textbf{13.71} \\ 
  \xgenwizard &
  \multicolumn{1}{c}{25.56} &
  \multicolumn{1}{c}{6.71} &
  16.84 &
  \multicolumn{1}{c}{\phantom{0}8.97} &
  \multicolumn{1}{c}{0.90} &
  \phantom{0}5.49 &
  \multicolumn{1}{c}{19.15} &
  \multicolumn{1}{c}{1.86} &
  \phantom{0}9.53 \\ 
   \vicuna-7B-v1.3 &
  \multicolumn{1}{c}{14.23} &
  \multicolumn{1}{c}{2.01} & \phantom{0}9.05
   &
  \multicolumn{1}{c}{16.49} &
  \multicolumn{1}{c}{1.00} & \phantom{0}9.99
   &
  \multicolumn{1}{c}{17.06} &
  \multicolumn{1}{c}{1.49} & \phantom{0}8.85
   \\ 
Falcon-7B-instruct &
  \multicolumn{1}{c}{14.89} &
  \multicolumn{1}{c}{1.97} &
  \phantom{0}9.28 &
  \multicolumn{1}{c}{18.90} &
  \multicolumn{1}{c}{1.80} &
  \phantom{0}9.37 &
  \multicolumn{1}{c}{18.90} &
  \multicolumn{1}{c}{1.80} &
  \phantom{0}9.37 \\ 
MPT-7B-instruct &
  \multicolumn{1}{c}{11.95} &
  \multicolumn{1}{c}{1.88} &
  \phantom{0}8.10 &
  \multicolumn{1}{c}{14.27} &
  \multicolumn{1}{c}{1.40} &
  \phantom{0}8.89 &
  \multicolumn{1}{c}{19.80} &
  \multicolumn{1}{c}{2.39} &
  10.23 \\ 
Alpaca-7B &
  \multicolumn{1}{c}{\phantom{0}9.69} &
  \multicolumn{1}{c}{1.77} &
  \phantom{0}6.43 &
  \multicolumn{1}{c}{16.26} &
  \multicolumn{1}{c}{1.56} &
  10.66 &
  \multicolumn{1}{c}{12.26} &
  \multicolumn{1}{c}{1.15} &
  \phantom{0}7.30 \\ 
  \wizardlm-7B &
  \multicolumn{1}{c}{18.97} &
  \multicolumn{1}{c}{2.65} & 10.32
   &
  \multicolumn{1}{c}{14.13} &
  \multicolumn{1}{c}{1.11} & \phantom{0}8.07
   &
  \multicolumn{1}{c}{19.16} &
  \multicolumn{1}{c}{1.87} & \phantom{0}9.51
   \\ 
  \bottomrule
\end{tabular}
\end{table}

\section{Carbon Footprint}

\begin{figure}[t]
  \centering
  \includegraphics[width=0.7\textwidth]{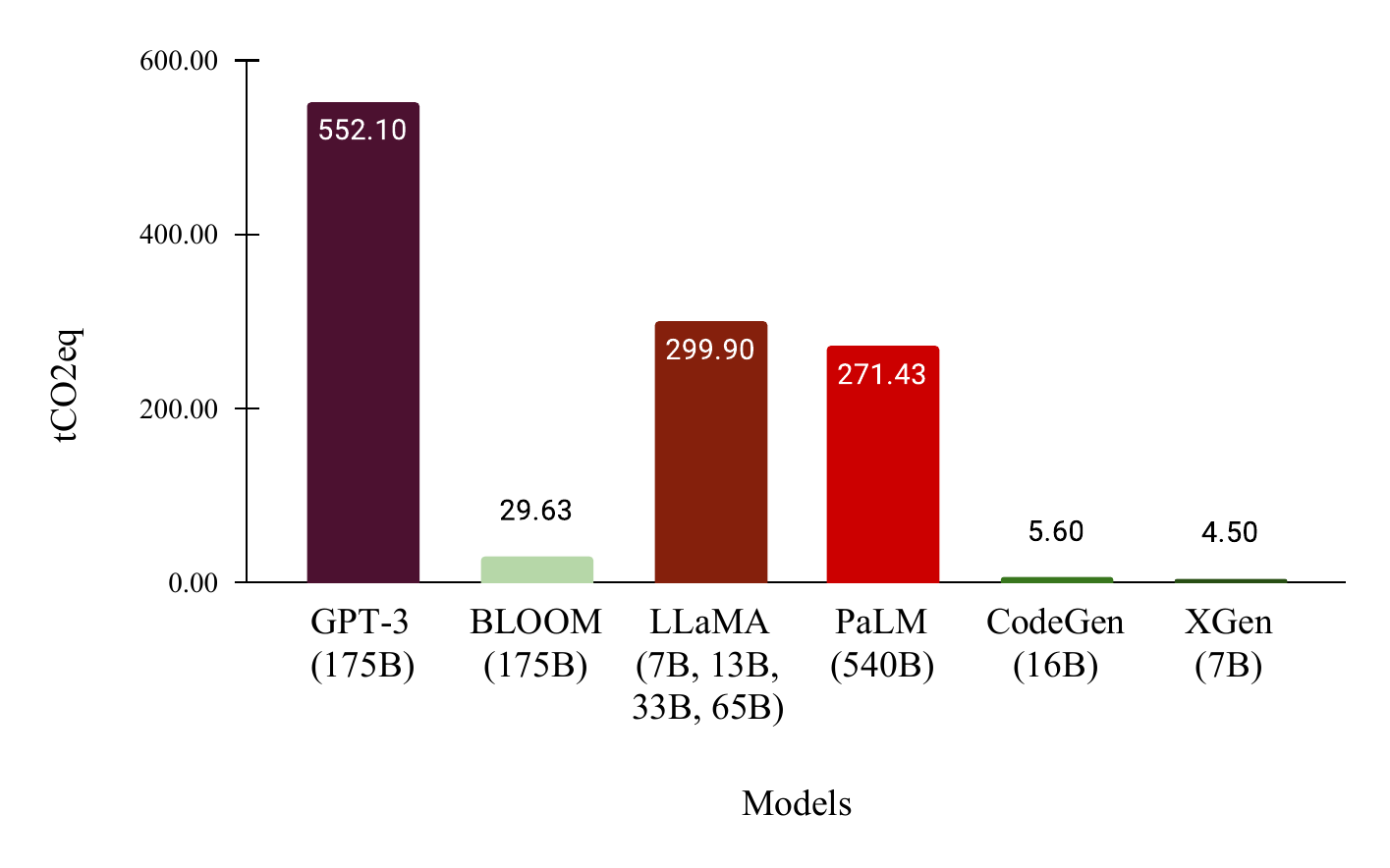}
  \caption{Carbon emission of different models.}
  \label{fig:carbon}
\end{figure}

To estimate the energy consumption and the resulting emission of carbon dioxide for training \xgen, we follow \cite{Wu-22}. Specifically, we compute Mega-watt-hour (MWh) as follows:  

\begin{eqnarray}
\text{MWh} &=& \text{TPU-hours} \times \text{(TPU power consumption)} \times \text{PUE} \\
        &=& 270,336 \times 192  \times 1.10 \\
        &=& 57
\end{eqnarray}

\noindent where we set the Power Usage Effectiveness (PUE)
to $1.10$ following the standard. The resulting carbon emission depends on the data center location. For \xgen, this amounts to:  $\text{tCO}_2\text{eq = MWh} ~(57)  \times 0.079 = 4.5$. In Figure \ref{fig:carbon}, we show this in comparison with other LLMs. 

\section{Note on Potential Risks}

Finally, despite our efforts in addressing the risks of bias, toxicity, and hallucinations both in pre-training and fine-tuning stages, like other LLMs, \xgen\ models are not free from such limitations. We hope our open-sourced codebase will help other researchers better understand these challenges and improve on these key limitations for making AI beneficial for everyone.

\section{Conclusion}

In this report, we have presented our newly developed \xgen\ models that support up to 8K tokens as input context. We described its effective stage-wise pre-training process with different sequence lengths (2K $\rightarrow$ 4K $\rightarrow$ 8K) and data mixtures (mostly text $\rightarrow$ 50\% text - 50\% code). We have shown that the resulting model achieves comparable or better results on standard text and code generation benchmarks compared to state-of-the-art open-source LLMs. 

We have also described the finetuning process of the \xgen\ model on two different public-domain instructional datasets, creating the \xgen-Inst counterparts. The results on two popular benchmarks show that our models often outperform existing models of similar sizes and sometimes even much larger models. We then evaluated the models on long sequence modeling tasks, which validates the superiority of our 8K-sequence model over the existing 2K-sequence LLMs. Finally, we hope that the open-sourcing of our models will contribute to open science in understanding the strengths and limitations of LLMs and will have significant impacts on business and commerce.  

\section{Author Contribution}

\small
\paragraph{Pre-training Model}
Erik Nijkamp (\textit{lead}), Hiroaki Hayashi, Tian Xie, Chen Xing
\paragraph{Pre-training Data}
Tian Xie (\textit{lead}), Hiroaki Hayashi, Lidiya Murakhovs’ka
\paragraph{Evaluation}
Congying Xia (\textit{lead}), Tian Xie, Erik Nijkamp, Rui Meng, Hiroaki Hayashi, Wojciech Kry{\'s}ci{\'n}ski, Ye Liu, Lifu Tu, Meghana Bhat
\paragraph{Instruction Tuning (Model)}
Bo Pang (\textit{lead}), Chen Xing

\paragraph{Instruction Tuning (Tool \& Data)} Jesse Vig,  Semih Yavuz, Chen Xing, Philippe Laban, Ben Krause, Senthil Purushwalkam, Tong Niu, Wojciech Kry{\'s}ci{\'n}ski, Lidiya Murakhovs'ka, Prafulla Kumar Choubey, Alex Fabbri, Ye Liu, Rui Meng,  Lifu Tu, Meghana Bhat

\paragraph{Project Coordination} Caiming Xiong (\textit{co-lead}), Shafiq Joty (\textit{co-lead}), Yingbo Zhou (\textit{co-lead}),  Chien-Sheng Wu, Silvio Savarese

\bibliography{anthology}

\begin{thebibliography}{10}

\bibitem{bisk2020piqa}
Yonatan Bisk, Rowan Zellers, Jianfeng Gao, Yejin Choi, et~al.
\newblock Piqa: Reasoning about physical commonsense in natural language.
\newblock In {\em Proceedings of the AAAI conference on artificial intelligence}, volume~34, pages 7432--7439, 2020.

\bibitem{brown2020language}
Tom Brown, Benjamin Mann, Nick Ryder, Melanie Subbiah, Jared~D Kaplan, Prafulla Dhariwal, Arvind Neelakantan, Pranav Shyam, Girish Sastry, Amanda Askell, et~al.
\newblock Language models are few-shot learners.
\newblock {\em Advances in neural information processing systems}, 33:1877--1901, 2020.

\bibitem{AMI}
Jean Carletta, Simone Ashby, Sebastien Bourban, Mike Flynn, Mael Guillemot, Thomas Hain, Jaroslav Kadlec, Vasilis Karaiskos, Wessel Kraaij, Melissa Kronenthal, Guillaume Lathoud, Mike Lincoln, Agnes Lisowska, Iain McCowan, Wilfried Post, Dennis Reidsma, and Pierre Wellner.
\newblock The ami meeting corpus: A pre-announcement.
\newblock In {\em Proceedings of the Second International Conference on Machine Learning for Multimodal Interaction}, MLMI'05, page 28–39, Berlin, Heidelberg, 2005. Springer-Verlag.

\bibitem{chen2021codex}
Mark Chen, Jerry Tworek, Heewoo Jun, Qiming Yuan, Henrique Ponde de~Oliveira Pinto, Jared Kaplan, Harri Edwards, Yuri Burda, Nicholas Joseph, Greg Brockman, et~al.
\newblock Evaluating large language models trained on code.
\newblock {\em arXiv preprint arXiv:2107.03374}, 2021.

\bibitem{chen-etal-2022-summscreen}
Mingda Chen, Zewei Chu, Sam Wiseman, and Kevin Gimpel.
\newblock {S}umm{S}creen: A dataset for abstractive screenplay summarization.
\newblock In {\em Proceedings of the 60th Annual Meeting of the Association for Computational Linguistics (Volume 1: Long Papers)}, pages 8602--8615, Dublin, Ireland, May 2022. Association for Computational Linguistics.

\bibitem{chowdhery2022palm}
Aakanksha Chowdhery, Sharan Narang, Jacob Devlin, Maarten Bosma, Gaurav Mishra, Adam Roberts, Paul Barham, Hyung~Won Chung, Charles Sutton, Sebastian Gehrmann, et~al.
\newblock Palm: Scaling language modeling with pathways.
\newblock {\em arXiv preprint arXiv:2204.02311}, 2022.

\bibitem{clark2019boolq}
Christopher Clark, Kenton Lee, Ming-Wei Chang, Tom Kwiatkowski, Michael Collins, and Kristina Toutanova.
\newblock Boolq: Exploring the surprising difficulty of natural yes/no questions.
\newblock {\em arXiv preprint arXiv:1905.10044}, 2019.

\bibitem{clark2018think}
Peter Clark, Isaac Cowhey, Oren Etzioni, Tushar Khot, Ashish Sabharwal, Carissa Schoenick, and Oyvind Tafjord.
\newblock Think you have solved question answering? try arc, the ai2 reasoning challenge.
\newblock {\em arXiv preprint arXiv:1803.05457}, 2018.

\bibitem{together2023redpajama}
Together Computer.
\newblock Redpajama: An open source recipe to reproduce llama training dataset, April 2023.

\bibitem{DatabricksBlog2023DollyV2}
Mike Conover, Matt Hayes, Ankit Mathur, Jianwei Xie, Jun Wan, Sam Shah, Ali Ghodsi, Patrick Wendell, Matei Zaharia, and Reynold Xin.
\newblock Free dolly: Introducing the world's first truly open instruction-tuned llm, 2023.

\bibitem{dubois2023alpacafarm}
Yann Dubois, Xuechen Li, Rohan Taori, Tianyi Zhang, Ishaan Gulrajani, Jimmy Ba, Carlos Guestrin, Percy Liang, and Tatsunori~B. Hashimoto.
\newblock Alpacafarm: A simulation framework for methods that learn from human feedback, 2023.

\bibitem{eval-harness}
Leo Gao, Jonathan Tow, Stella Biderman, Sid Black, Anthony DiPofi, Charles Foster, Laurence Golding, Jeffrey Hsu, Kyle McDonell, Niklas Muennighoff, Jason Phang, Laria Reynolds, Eric Tang, Anish Thite, Ben Wang, Kevin Wang, and Andy Zou.
\newblock A framework for few-shot language model evaluation, September 2021.

\bibitem{hendrycks2020measuring}
Dan Hendrycks, Collin Burns, Steven Basart, Andy Zou, Mantas Mazeika, Dawn Song, and Jacob Steinhardt.
\newblock Measuring {M}assive {M}ultitask {L}anguage {U}nderstanding.
\newblock In {\em International Conference on Learning Representations}, 2021.

\bibitem{hoffmann2022an}
Jordan Hoffmann, Sebastian Borgeaud, Arthur Mensch, Elena Buchatskaya, Trevor Cai, Eliza Rutherford, Diego de~las Casas, Lisa~Anne Hendricks, Johannes Welbl, Aidan Clark, Tom Hennigan, Eric Noland, Katherine Millican, George van~den Driessche, Bogdan Damoc, Aurelia Guy, Simon Osindero, Karen Simonyan, Erich Elsen, Oriol Vinyals, Jack~William Rae, and Laurent Sifre.
\newblock An empirical analysis of compute-optimal large language model training.
\newblock In Alice~H. Oh, Alekh Agarwal, Danielle Belgrave, and Kyunghyun Cho, editors, {\em Advances in Neural Information Processing Systems}, 2022.

\bibitem{li2023starcoder}
Raymond Li, Loubna~Ben Allal, Yangtian Zi, Niklas Muennighoff, Denis Kocetkov, Chenghao Mou, Marc Marone, Christopher Akiki, Jia Li, Jenny Chim, Qian Liu, Evgenii Zheltonozhskii, Terry~Yue Zhuo, Thomas Wang, Olivier Dehaene, Mishig Davaadorj, Joel Lamy-Poirier, João Monteiro, Oleh Shliazhko, Nicolas Gontier, Nicholas Meade, Armel Zebaze, Ming-Ho Yee, Logesh~Kumar Umapathi, Jian Zhu, Benjamin Lipkin, Muhtasham Oblokulov, Zhiruo Wang, Rudra Murthy, Jason Stillerman, Siva~Sankalp Patel, Dmitry Abulkhanov, Marco Zocca, Manan Dey, Zhihan Zhang, Nour Fahmy, Urvashi Bhattacharyya, Wenhao Yu, Swayam Singh, Sasha Luccioni, Paulo Villegas, Maxim Kunakov, Fedor Zhdanov, Manuel Romero, Tony Lee, Nadav Timor, Jennifer Ding, Claire Schlesinger, Hailey Schoelkopf, Jan Ebert, Tri Dao, Mayank Mishra, Alex Gu, Jennifer Robinson, Carolyn~Jane Anderson, Brendan Dolan-Gavitt, Danish Contractor, Siva Reddy, Daniel Fried, Dzmitry Bahdanau, Yacine Jernite, Carlos~Muñoz Ferrandis, Sean Hughes, Thomas Wolf, Arjun Guha, Leandro von
  Werra, and Harm de~Vries.
\newblock Starcoder: may the source be with you!, 2023.

\bibitem{alpaca_eval}
Xuechen Li, Tianyi Zhang, Yann Dubois, Rohan Taori, Ishaan Gulrajani, Carlos Guestrin, Percy Liang, and Tatsunori~B. Hashimoto.
\newblock Alpacaeval: An automatic evaluator of instruction-following models.
\newblock \url{https://github.com/tatsu-lab/alpaca_eval}, 2023.

\bibitem{lin2021truthfulqa}
Stephanie Lin, Jacob Hilton, and Owain Evans.
\newblock Truthfulqa: Measuring how models mimic human falsehoods.
\newblock {\em arXiv preprint arXiv:2109.07958}, 2021.

\bibitem{longpre2023flan}
Shayne Longpre, Le~Hou, Tu~Vu, Albert Webson, Hyung~Won Chung, Yi~Tay, Denny Zhou, Quoc~V Le, Barret Zoph, Jason Wei, et~al.
\newblock The flan collection: Designing data and methods for effective instruction tuning.
\newblock {\em arXiv preprint arXiv:2301.13688}, 2023.

\bibitem{mihaylov2018can}
Todor Mihaylov, Peter Clark, Tushar Khot, and Ashish Sabharwal.
\newblock Can a suit of armor conduct electricity? a new dataset for open book question answering.
\newblock {\em arXiv preprint arXiv:1809.02789}, 2018.

\bibitem{molybog2023theory}
Igor Molybog, Peter Albert, Moya Chen, Zachary DeVito, David Esiobu, Naman Goyal, Punit~Singh Koura, Sharan Narang, Andrew Poulton, Ruan Silva, et~al.
\newblock A theory on adam instability in large-scale machine learning.
\newblock {\em arXiv preprint arXiv:2304.09871}, 2023.

\bibitem{nicholson2020bard}
Ann~E Nicholson, Kevin~B Korb, Erik~P Nyberg, Michael Wybrow, Ingrid Zukerman, Steven Mascaro, Shreshth Thakur, Abraham~Oshni Alvandi, Jeff Riley, Ross Pearson, et~al.
\newblock Bard: A structured technique for group elicitation of bayesian networks to support analytic reasoning.
\newblock {\em arXiv preprint arXiv:2003.01207}, 2020.

\bibitem{Jaxformer}
Erik Nijkamp.
\newblock Jaxformer: A minimal library for training llms on tpu.
\newblock \url{https://github.com/salesforce/jaxformer}, 2022.

\bibitem{nijkamp2022codegen}
Erik Nijkamp, Bo~Pang, Hiroaki Hayashi, Lifu Tu, Huan Wang, Yingbo Zhou, Silvio Savarese, and Caiming Xiong.
\newblock Codegen: An open large language model for code with multi-turn program synthesis.
\newblock {\em arXiv preprint arXiv:2203.13474}, 2022.

\bibitem{openai2023gpt4}
OpenAI.
\newblock Gpt-4 technical report, 2023.

\bibitem{instructgpt_ouyang2022training}
Long Ouyang, Jeffrey Wu, Xu~Jiang, Diogo Almeida, Carroll Wainwright, Pamela Mishkin, Chong Zhang, Sandhini Agarwal, Katarina Slama, Alex Ray, et~al.
\newblock Training language models to follow instructions with human feedback.
\newblock {\em Advances in Neural Information Processing Systems}, 35:27730--27744, 2022.

\bibitem{press2022train}
Ofir Press, Noah Smith, and Mike Lewis.
\newblock Train short, test long: Attention with linear biases enables input length extrapolation.
\newblock In {\em International Conference on Learning Representations}, 2022.

\bibitem{sakaguchi2021winogrande}
Keisuke Sakaguchi, Ronan~Le Bras, Chandra Bhagavatula, and Yejin Choi.
\newblock Winogrande: An adversarial winograd schema challenge at scale.
\newblock {\em Communications of the ACM}, 64(9):99--106, 2021.

\bibitem{sanh2021multitask}
Victor Sanh, Albert Webson, Colin Raffel, Stephen~H Bach, Lintang Sutawika, Zaid Alyafeai, Antoine Chaffin, Arnaud Stiegler, Teven~Le Scao, Arun Raja, et~al.
\newblock Multitask prompted training enables zero-shot task generalization.
\newblock {\em arXiv preprint arXiv:2110.08207}, 2021.

\bibitem{sennrich2015neural}
Rico Sennrich, Barry Haddow, and Alexandra Birch.
\newblock Neural machine translation of rare words with subword units.
\newblock {\em arXiv preprint arXiv:1508.07909}, 2015.

\bibitem{shaham-etal-2022-scrolls}
Uri Shaham, Elad Segal, Maor Ivgi, Avia Efrat, Ori Yoran, Adi Haviv, Ankit Gupta, Wenhan Xiong, Mor Geva, Jonathan Berant, and Omer Levy.
\newblock {SCROLLS}: Standardized {C}ompa{R}ison over long language sequences.
\newblock In {\em Proceedings of the 2022 Conference on Empirical Methods in Natural Language Processing}, pages 12007--12021, Abu Dhabi, United Arab Emirates, December 2022. Association for Computational Linguistics.

\bibitem{shazeer2020glu}
Noam Shazeer.
\newblock Glu variants improve transformer.
\newblock {\em arXiv preprint arXiv:2002.05202}, 2020.

\bibitem{alpaca}
Rohan Taori, Ishaan Gulrajani, Tianyi Zhang, Yann Dubois, Xuechen Li, Carlos Guestrin, Percy Liang, and Tatsunori~B. Hashimoto.
\newblock Stanford alpaca: An instruction-following llama model.
\newblock \url{https://github.com/tatsu-lab/stanford_alpaca}, 2023.

\bibitem{tay2021long}
Yi~Tay, Mostafa Dehghani, Samira Abnar, Yikang Shen, Dara Bahri, Philip Pham, Jinfeng Rao, Liu Yang, Sebastian Ruder, and Donald Metzler.
\newblock Long range arena : A benchmark for efficient transformers.
\newblock In {\em International Conference on Learning Representations}, 2021.

\bibitem{touvron2023llama}
Hugo Touvron, Thibaut Lavril, Gautier Izacard, Xavier Martinet, Marie-Anne Lachaux, Timoth{\'e}e Lacroix, Baptiste Rozi{\`e}re, Naman Goyal, Eric Hambro, Faisal Azhar, et~al.
\newblock Llama: Open and efficient foundation language models.
\newblock {\em arXiv preprint arXiv:2302.13971}, 2023.

\bibitem{gpt-j}
Ben Wang and Aran Komatsuzaki.
\newblock {GPT-J-6B: A 6 Billion Parameter Autoregressive Language Model}.
\newblock \url{https://github.com/kingoflolz/mesh-transformer-jax}, May 2021.

\bibitem{supernaturalinstructions}
Yizhong Wang, Swaroop Mishra, Pegah Alipoormolabashi, Yeganeh Kordi, Amirreza Mirzaei, Anjana Arunkumar, Arjun Ashok, Arut~Selvan Dhanasekaran, Atharva Naik, David Stap, et~al.
\newblock Super-naturalinstructions:generalization via declarative instructions on 1600+ tasks.
\newblock In {\em EMNLP}, 2022.

\bibitem{wei2021finetuned}
Jason Wei, Maarten Bosma, Vincent~Y Zhao, Kelvin Guu, Adams~Wei Yu, Brian Lester, Nan Du, Andrew~M Dai, and Quoc~V Le.
\newblock Finetuned language models are zero-shot learners.
\newblock {\em arXiv preprint arXiv:2109.01652}, 2021.

\bibitem{Wu-22}
Carole-Jean Wu, Ramya Raghavendra, Udit Gupta, Bilge Acun, Newsha Ardalani, Kiwan Maeng, Gloria Chang, Fiona Aga, Jinshi Huang, Charles Bai, Michael Gschwind, Anurag Gupta, Myle Ott, Anastasia Melnikov, Salvatore Candido, David Brooks, Geeta Chauhan, Benjamin Lee, Hsien-Hsin Lee, Bugra Akyildiz, Maximilian Balandat, Joe Spisak, Ravi Jain, Mike Rabbat, and Kim Hazelwood.
\newblock Sustainable ai: Environmental implications, challenges and opportunities.
\newblock In D.~Marculescu, Y.~Chi, and C.~Wu, editors, {\em Proceedings of Machine Learning and Systems}, volume~4, pages 795--813, 2022.

\bibitem{xu2023wizardlm}
Can Xu, Qingfeng Sun, Kai Zheng, Xiubo Geng, Pu~Zhao, Jiazhan Feng, Chongyang Tao, and Daxin Jiang.
\newblock Wizardlm: Empowering large language models to follow complex instructions.
\newblock {\em arXiv preprint arXiv:2304.12244}, 2023.

\bibitem{xu2023baize}
Canwen Xu, Daya Guo, Nan Duan, and Julian McAuley.
\newblock Baize: An open-source chat model with parameter-efficient tuning on self-chat data.
\newblock {\em arXiv preprint arXiv:2304.01196}, 2023.

\bibitem{zellers2019hellaswag}
Rowan Zellers, Ari Holtzman, Yonatan Bisk, Ali Farhadi, and Yejin Choi.
\newblock Hellaswag: Can a machine really finish your sentence?
\newblock {\em arXiv preprint arXiv:1905.07830}, 2019.

\bibitem{zhang2019root}
Biao Zhang and Rico Sennrich.
\newblock Root mean square layer normalization.
\newblock {\em Advances in Neural Information Processing Systems}, 32, 2019.

\bibitem{zheng2023judging}
Lianmin Zheng, Wei-Lin Chiang, Ying Sheng, Siyuan Zhuang, Zhanghao Wu, Yonghao Zhuang, Zi~Lin, Zhuohan Li, Dacheng Li, Eric.~P Xing, Hao Zhang, Joseph~E. Gonzalez, and Ion Stoica.
\newblock Judging llm-as-a-judge with mt-bench and chatbot arena, 2023.

\end{thebibliography}
\bibliographystyle{plain}


\newpage
\appendix
\section{Appendix}
\label{appendix}
\subsection{Long form QA - Prompts used for Question Generation}
\label{app:question_generation}
We formulate our question generation method as a two-step process: (1) Summarization and (2) Question generation from summary. 
In the first step, we design a prompt for generating a summary as shown below: \\

{\textit{Summarize the paragraphs below in the context of \{title\} in \{domain\}.}} \\

In the next step, we ask ChatGPT to generate questions from summary as shown below: \\

{\textit{Using the context below, come up with follow-up questions such that answers are beyond few words or a couple of phrases. Rank the generated questions in the order of decreasing complexity to answer and display only the top 3. 
\{context\}}}

To demonstrate the usefulness of our question generation process, we also establish a baseline with the same instructions where questions are directly generated from the passage. The prompt used for the baseline is:\\

{ \textit{Using the context below, come up with follow-up questions such that answers are beyond few words or a couple of phrases. Rank the generated questions in the order of decreasing complexity to answer and display only the top 3. 
\{context\}}} \\
\end{document}